# Disjoint Multi-task Learning between Heterogeneous Human-centric Tasks

Dong-Jin Kim [1]   Jinsoo Choi[1]   Tae-Hyun Oh[2]   Youngjin Yoon[1]   In So Kweon[1]

[1]KAIST, Daejeon, South Korea.
[2]MIT CSAIL, Cambridge, MA.

## Abstract

*Human behavior understanding is arguably one of the most important mid-level components in artificial intelligence. In order to efficiently make use of data, multi-task learning has been studied in diverse computer vision tasks including human behavior understanding. However, multi-task learning relies on task specific datasets and constructing such datasets can be cumbersome. It requires huge amounts of data, labeling efforts, statistical consideration etc. In this paper, we leverage existing single-task datasets for human action classification and captioning data for efficient human behavior learning. Since the data in each dataset has respective heterogeneous annotations, traditional multi-task learning is not effective in this scenario. To this end, we propose a novel alternating directional optimization method to efficiently learn from the heterogeneous data. We demonstrate the effectiveness of our model and show performance improvements on both classification and sentence retrieval tasks in comparison to the models trained on each of the single-task datasets.*

## 1. Introduction

One of the basic artificial intelligence (AI) components of fundamental importance would be human behavior understanding, in that there are many human centric visual tasks which typically require a certain level of human behavior understanding, *e.g*., learning from demonstration (imitation learning) [2, 21, 38], video captioning [53] . The human-centric tasks may benefit from mid-level understandings such as human detection [33], pose detection [7, 19, 49] , action classifications [24, 41, 43], human-object interactions [13, 18, 51], *etc*. These are getting realized by virtue of recent advances in deep neural networks (DNN). This typically requires a large amount of training data; hence, as more data is leveraged, it is expected to produce better performance. A common way to boost data ef-ficiency is multi-task learning that shares a common knowledge for multiple tasks, *i.e*., multi-task learning (MTL), or improving the performances of individual tasks.

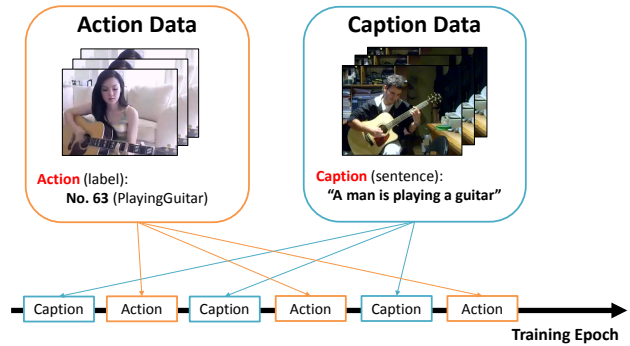

Figure 1. The description of the proposed alternative directional optimization method for training heterogeneous classification and captioning data. The training datasets have no intersection.

In this work, we postulate that the human-centric tasks are based on a common human behavior understanding; hence, sharing human behavior information between multiple tasks can enhance the performance of human-centric task systems. Among them, we focus on the action classification and captioning due to two reasons: 1) their labels mainly stem from the human's behaviors, and 2) while they are contextually similar, the tasks require different levels of understanding, *i.e*., holistic action class understanding vs. human and object interaction. The latter notion often refers to as hierarchical knowledge, [37], which may help both levels to easily find good knowledge from each other. In this paper, we verify the hypothesis with several experiments.

Comparing to single task learning, the MTL may be regarded as a way to use data effectively, but deep learning based MTL still requires not only large scale data but also multi-task labels per single data; *i.e*., we need a large scale data that is specifically designed for multi-task. However, constructing a new large-scale multi-task dataset can be cumbersome and time-consuming. We jointly utilize exist-ing heterogeneous single-task datasets, so that we can avoid the challenge of data collection while leveraging to complement each other in the form of the MTL.

It is not trivial to train a multi-task network with datasets of which data only has a label for either task, not both. We call this training setup as *disjoint multi-task learning* (DML). A naive approach for this would be to train a network in a way that alternates training batches from either of the task datasets and back-propagate through the output branches corresponding to the task. However, due to the well-known forgetting effect [29], this naive approach easily loses the learned knowledge of a task, when we back-propagate through the output branch of the other task. Intuitively, when training task A, the back-propagation fits the network parameters of the shared part and the output branch of the task A to the task A, whereby the parameters of the other task B remain the same and turn out to be incompatible with the shared part. In order to prevent such repetition of learning and forgetting, we preserve knowledge of a task while training for the other task.

The contributions of this work are as follows.
1. We jointly learn both action classification and captioning data to prevent forgetting effects to outperform the single-task and multi-task baselines.
2. We propose an effective method for training the multi-task convolutional neural network (CNN) model with heterogeneous datasets with different tasks.
3. We systematically analyze the proposed method in various perspectives, qualitatively and quantitatively.

## 2. Related Work

Previous works extend over multiple contexts: human understanding, multi-task learning and disjoint setups. We briefly review the relevant works in a categorized way to show where our work stands in different perspectives.

**Leveraging human property** A representative application of leveraging the presence of human would be action recognition. CNN based approaches have achieved impressive performances [8, 14, 15, 23, 24, 41, 43, 47, 48]. Since human action information typically presents across time, various architectures have been proposed to capture structure of action: Karpathy *et al*. [24] use multiple CNN concatenation across time, 3D convolution [43] (C3D) operates across local spatial and temporal axis, and two stream networks [41, 47] leverage explicit optical flow as another input modality. In this work, we build our architecture on top of C3D for video inputs, which does not require expensive optical flow estimation.

Besides action recognition, since understanding levels of human action (*i.e.*, human behavior) are all different depending on tasks, there have been various definitions of human action. Herath *et al*. [20] suggest the definition of *action* is "the most elementary human-surrounding interaction with a meaning."[1] Basically, explicit human-object interaction modeling [13, 18, 51] has shown up improvement of recognition performance. However, they require predetermined classes of relational information that may not deal with undefined classes; hence, we do not explicitly restrict them. On the other hand, image captioning tasks deal with semantic representations and understanding of images which do not require predetermined object classes. Learning between the image and caption modalities enable using the rich semantic information [26, 45, 50, 52].

Since captioning task is designed to describe visual information based on perfect understanding, captioning models provide implicit ways to understand humans and surrounding objects. Its distilled information may differ from the action recognition task. Thus, we postulate that respective cues from action recognition and captioning tasks compensate each other. We learn a CNN network in multiple perspectives of tasks, *i.e.*, transfer learning and multi-task learning, so that the model implicitly deals with the surrounding information of the human, but without any external information such as skeletons or bounding boxes [13, 18, 32, 51].

**Transfer/multi-task learning** Training a model with multiple task labels is broadly used either to overcome the lack of training data or to improve the training performance. Oquab *et al*. [34] propose deep transfer learning, which fine-tunes a network pre-trained on the large scale ImageNet dataset [36] to a new target task. Another typical way of joint learning for multiple tasks is Multi-task learning (MTL) [9]. By coupling the MTL with deep learning, shared lower-layers of a model are learned to be generalized to multiple tasks, which reduces the burden for learning task-specific branches stemming from the shared part. This MTL has shown good regularization effect and performance improvement in diverse computer vision tasks [4, 5, 12, 17, 32, 54, 55].

**Multi-task learning with disjoint datasets** We often have disjoint datasets that do not have intersection of training data and label modalities in two sets. There are a few works on disjoint datasets in machine translation [31], action grouping [30], universal computer vision network [27] (from low- to high-level computer vision tasks), multi-task feature learning [46], and universal representations [6]. Most of these methods update each branch of the model alternately in a naive transfer learning way. Since both transfer and multi-task learning schemes suffer from the forgetting effect [29], they exploit lots of large scale data. We show that such a naive alternating training is not efficient and even degrades performance in multi-task learning regime.

---
[1] For thorough survey of modern action understanding, one can refer to Herath *et al*.

In order to address the forgetting problem, there have been several methods such as learning without forgetting [29], lifelong learning [1, 40] and continual learning [25, 29, 39] methods, which are methods to train multiple tasks sequentially. However, these methods are for leveraging source task information to obtain improvements in the target tasks, whereas our goal is to give benefits to either or both of the tasks.

We extend the transfer learning method of Li *et al.* [29] for training with the disjoint multi-task setup, so that both tasks benefit each other during training and lead to faster convergence as well as better performance. This scheme does not require multi-task labels for training inputs in contrast to the MTL.

## 3. Disjoint Multi-task Learning

In this work, we hypothesize that captioning data may be helpful for action classification for two reasons. First, two tasks are contextually similar as human-centric tasks. If we compare the videos in UCF101 action recognition data [42] and YouTube2Text captioning data [11], the contents are mostly about human actions. Second, sentences have richer information than a simple label. In one sentence, there is information about not only the class of the data but also general semantic relationships that describe the essential contents. Therefore, we believe that captioning data might be useful for multi-task learning with a classification dataset.

To validate the hypothesis, we use CNN model as a shared network and we add two task-specific branches to be multi-task learning of classification and caption semantic embedding. The description of our multi-task model is depicted in Figure 2.

### 3.1. A Baseline for Disjoint Multi-task Learning

We deal with a multi-task problem consisting of classification and semantic embedding. Let us denote a video data as $v \in \mathcal{V}$. Given an input video $v$, the output of the classification model $f_A$ is a $K$-dimensional softmax probability vector $\hat{y}_A$, which is learned from the ground truth action label $y_A$. For this task, we use the typical cross-entropy loss:

$$\mathcal{L}_{cls}(y_A, \hat{y}_A) = -\sum_{k=1}^{K} y_A^k \log \hat{y}_A^k. \qquad (1)$$

For the sentence embedding, we first embed the ground truth sentences with the state-of-the-art pre-trained semantic embedding model [44]. These embedding vectors are considered as ground truth sentence embedding vectors $y_S$. The sentence embedding branch infers a unit vector $\hat{y}_S$ learned from embedding vectors $y_S$ of the ground truth sentences. We use the cosine distance loss between the ground truth embedding and the predicted embedding vector.

$$\mathcal{L}_{emb}(y_S, \hat{y}_S) = -y_S \cdot \hat{y}_S. \qquad (2)$$

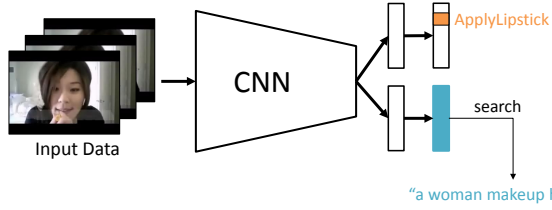

Figure 2. Our multi-task convolutional neural network (CNN). The model is based on CNN with two task branches for classification and caption embedding respectively.

Combining these two task-specific losses with the weighted sum of $\lambda$, we have the following optimization problem:

$$\min_{\{\theta.\}} \sum_{v \in \mathcal{V}} \lambda \mathcal{L}_{cls}(y_A, f_A(\theta_R, \theta_A, v)) \\ + (1-\lambda)\mathcal{L}_{emb}(y_S, f_S(\theta_R, \theta_S, v)), \qquad (3)$$

where $\theta_R, \theta_A$ and $\theta_S$ represent model weight parameters for the shared root network, action branch, and sentence branch respectively, and $\lambda$ is a multi-task parameter.

In a typical multi-task learning scenario, one may try to train the model by conventional multi-task back propagation, where the model back propagates gradients from both ends of branches. This can be depicted as follows:

$$\min_{\{\theta.\}} \sum_{v_A \in \mathcal{V}_A} \lambda_A \mathcal{L}_{cls}(y_{AA}, f_A(\theta_R, \theta_A, v_A)) \\ + (1-\lambda_A)\mathcal{L}_{emb}(y_{SA}, f_S(\theta_R, \theta_S, v_A)) \\ + \sum_{v_S \in \mathcal{V}_S} \lambda_S \mathcal{L}_{cls}(y_{AS}, f_A(\theta_R, \theta_A, v_S)) \\ + (1-\lambda_S)\mathcal{L}_{emb}(y_{SS}, f_S(\theta_R, \theta_S, v_S)), \qquad (4)$$

where $y_{AA}$ and $y_{AC}$ are action and caption label respectively for action classification data $\mathcal{V}_A$, and $y_{AS}$ and $y_{SS}$ are for caption data $\mathcal{V}_S$.

However, there is no way to directly train the objective loss in Eq. (4) by the multi-task back propagation because each input video has only either task label. Namely, separately considering videos in an action classification dataset, *i.e.*, $v_A \in \mathcal{V}_A$, and in a caption dataset, *i.e.*, $v_S \in \mathcal{V}_S$, a video $v_A$ from the classification dataset has no corresponding ground truth data $y_{SC}$ and vice versa for the caption dataset. This is the key problem we wanted to solve. We define this learning scenario as DML and address an appropriate optimization method to solve this problem.

A naive approach is an alternating learning for each branch at a time. Specifically, suppose that the training starts from the caption dataset. The shared network and caption branch of the model can be first trained with the caption dataset based only on $\mathcal{L}_{emb}$ in Eq. (3) by setting $\mathcal{L}_{cls} = 0$. After one epoch of training on captioning dataset is done, in this round, the model starts training on a classification dataset with respect to $\mathcal{L}_{cls}$ in Eq. (3) by setting $\mathcal{L}_{emb} = 0$. This procedure is iteratively applied to the end. The total loss function can be depicted as follows:

$$\min_{\{\theta.\}} \sum_{v_A \in \mathcal{V}_A} \lambda_A \mathcal{L}_{cls}(y_{AA}, f_A(\theta_R, \theta_A, v_A)) \\ + \sum_{v_S \in \mathcal{V}_S} (1 - \lambda_S) \mathcal{L}_{emb}(y_{SS}, f_S(\theta_R, \theta_S, v_S)). \quad (5)$$

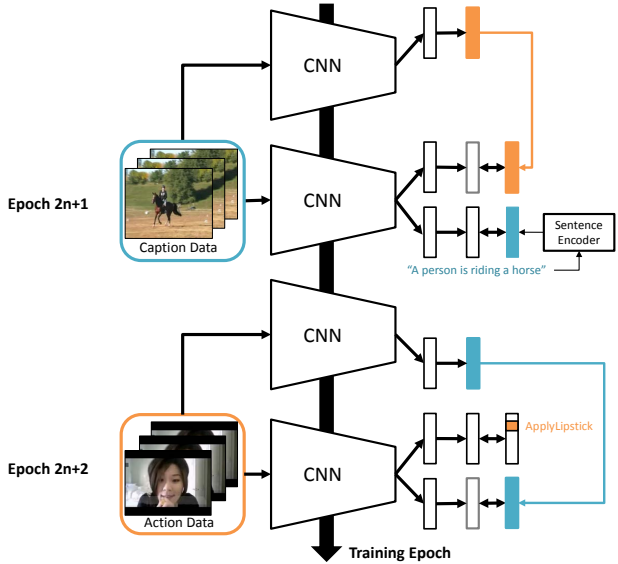

Figure 3. Our training procedure. The first data fed to the model is from the captioning data. Input data from each task is fed to the model and the model is updated with respect to the respective losses for each task. With our method, by reducing forgetting effect for alternating learning method, we facilitate the disjoint multi-task learning with single-task datasets.

The loss consists of classification and caption related losses. Each loss is alternately optimized.

Unfortunately, there is a well-known issue with this simple method. When we train either branch with a dataset, the knowledge of another task will be forgotten [29]. It is because during training a task, the optimization path of the shared network can be independent to one of the other task. Thus, the network would easily forget trained knowledge from the other task at every epoch, and optimizing the total loss in Eq. (5) is not likely to be converged. Therefore, while training without preventing this forgetting effect, the model repeats forgetting each of the tasks, whereby the model receives disadvantages compared to training with single data.

### 3.2. Dealing with Forgetting Effect

In order to solve the forgetting problem of alternating learning, we exploit a transfer learning method between multiple datasets called "Learning without Forgetting (LWF)" [29] which has been originally proposed to preserve the original information. The hypothesis is that the activation from the previous model contains the information of the source data and preserving it makes the model remember the information. Using this, we prevent forgetting during our alternating optimization. In order to prevent the forgetting effect, we utilize the "Knowledge distillation loss" [22] for preserving the activation of the previous task as follows:

$$\mathcal{L}_{distill}(y_A, \hat{y}_A) = -\sum_{k=1}^{K} {y'}_A^k \log \hat{y'}_A^k, \quad (6)$$

$$y'^k_A = \frac{(y_A^k)^{1/T}}{\sum_k (y_A^k)^{1/T}}. \quad (7)$$

However, LWF method is different from our task. First, the method is for simple transfer learning task. In our alternating strategy, this loss function is used for preserving the information of the previous training step. Second, the method was originally proposed only for image classification task, and thus only tested on the condition with similar source and target image pairs, such as ImageNet and VOC datasets. In this work, we apply LWF method to action classification and semantic embedding pair.

### 3.3. Proposed Method

In order to apply LWF method to our task, a few modifications are required. For semantic embedding, we use cosine distance loss in Eq. (2) which is different from cross-entropy loss. Therefore, the condition is not the same as when they used knowledge distillation loss. Semantic embedding task does not deal with class probability, so we think knowledge distillation loss is not appropriate for caption activation. Instead, we use the distance based embedding loss $\mathcal{L}_{emb}$ for distilling caption activation. In addition, while [29] simply used 1.0 for multi-task coefficient $\lambda$ in Eq. (3), because of the difference between cross-entropy loss and distance loss, a proper value for $\lambda$ is required, and we set different $\lambda$ values for classification and caption data as follows:

$$\mathcal{L}_A = \lambda_A \mathcal{L}_{cls} + (1 - \lambda_A) \mathcal{L}_{emb}, \quad (8)$$

$$\mathcal{L}_S = \lambda_S \mathcal{L}_{distill} + (1 - \lambda_S) \mathcal{L}_{emb}, \quad (9)$$

where $\mathcal{L}_A$ and $\mathcal{L}_S$ are the loss functions for action classification data and caption data respectively. Therefore, our final network is updated based on the following optimization problem:

$$\min_{\{\theta.\}} \sum_{v_A \in \mathcal{V}_A} \lambda_A \mathcal{L}_{cls}(y_{AA}, f_A(\theta_R, \theta_A, v_A))$$
$$+ (1 - \lambda_A)\mathcal{L}_{emb}(\bar{y}_{SA}, f_S(\theta_R, \theta_S, v_A))$$
$$+ \sum_{v_S \in \mathcal{V}_S} \lambda_S \mathcal{L}_{distill}(\bar{y}_{AS}, f_A(\theta_R, \theta_A, v_S)) \quad (10)$$
$$+ (1 - \lambda_S)\mathcal{L}_{emb}(y_{SS}, f_S(\theta_R, \theta_S, v_S)),$$

where $\bar{y}_{SA}$ is the extracted activation from the last layer of the sentence branch from the action classification data and vice versa for $\bar{y}_{AS}$. Our idea is that, for multi-task learning scenario, we consider missing variables $\bar{y}_{SA}$ and $\bar{y}_{AS}$, which are unknown labels, as trainable variables. For every epoch, we are able to update both functions $f_A$ and $f_S$ by utilizing $\bar{y}_{SA}$ or $\bar{y}_{AS}$, while $\bar{y}_{SA}$ and $\bar{y}_{AS}$ are also updated based on new data while preserving the information of the old data.

Our final training procedure is illustrated in Figure 3. First, when captioning data is applied to the network, we extract the class prediction $\hat{y}$ corresponding to the input data and save the activations. The activation is used as a supervision for knowledge distillation loss parallel to the typical caption loss in order to facilitate multi-task learning so that the model would reproduce the activation similar to the activation from the previous parameter. Trained sentence representation in this step is used to collect activations when classification data is fed to the network in the next step. Same as the previous step, we can also facilitate multi-task learning for classification data.

When test video is applied, trained multi-task network is used to predict class and to extract caption embedding as depicted in Figure 2. With this caption embedding, we can search the nearest sentence from the candidates.

## 4. Experiments

We compare among four experimental groups. The first one is the model only trained on the classification dataset and the second one is a caption-only model. The last two methods are a naive alternating method without LWF method and our final method.

We conduct the first experiments on the action-caption disjoint setting, and then to verify the benefit of human centric disjoint tasks, we compare the former results with the results from image classification and caption disjoint setting. We also provide further empirical analysis of the proposed method.

### 4.1. Training Details

For video data, we use state-of-the-art 3D CNN model [43], which feeds 16 continuous clip of frames, pre-trained on Sports-1M [24] dataset as a shared network. For image data we use VGG-S model [10] pre-trained on ImageNet dataset [36]. For caption semantic embedding task, we use state-of-the-art image semantic embedding model [44] as a sentence encoder. We also add L2 normalization for the output embedding. We add a new fully-connected layer from the fc7 layer of the shared network as task-specific branches. Adam [3] algorithm, with learning rate $5e^{-5}$ and $1e^{-5}$ for image and video classification experiment respectively, is applied for fast convergence. We use a batch size of 16 for video input and 64 for image input.

|  | Action | | Caption | | | |
| --- | --- | --- | --- | --- | --- | --- |
|  | Hit@1 | Acc | R@1 | R@5 | Med r | Mean r |
| Action only | 76.83 | 80.99 | - | - | - | - |
| Caption only | - | - | 11.3 | 49.2 | 5.8 | 6.5 |
| DML (w/o LWF) | 75.76 | 78.64 | 10.7 | 47.9 | 5.9 | 6.5 |
| DML + LWF | **78.03** | **82.26** | **11.5** | **49.4** | **5.7** | **6.4** |

Table 1. Comparison results on UCF101 - YouTube2Text dataset pair. The proposed model outperforms both action-only model and caption-only model.

We use action and caption metrics to measure our performance. For action task, we use Hit@1 and accuracy, which are clip-level and video-level accuracy respectively. Higher for the both the better. For image task, we use mAP measure. For caption task, we use rank at k (denoted by R@k) which is sentence recall at top rank k, and Median and Mean rank. Higher the rank at k the better, and lower the rank the better. For video datasets, we use 1 and 5 for k, and for image dataset, we use 1, 5 and 10 for k.

### 4.2. Multi-task with Heterogeneous Video Data

As a video action recognition dataset, we use either UCF101 dataset [42] or HMDB51 [28] dataset, which are the most popular action recognition datasets. UCF101 dataset consists of totally 13320 videos with average length 7.2 seconds, and human action labels with 101 classes. HMDB51 dataset contains totally 6766 videos of action labels with 51 classes. For caption dataset, we use YouTube2Text dataset [11], which was proposed for video captioning task. The dataset has 1970 videos (1300 for training, 670 for test) crawled from YouTube. Each of the video clips is around 10 seconds long and labeled with around 40 sentences of video descriptions in English (totally 80827 sentences.). In this paper, we collect 16-frames video clip with subsampling ratio 3. For UCF101 dataset, we collect video clips with 150 frames interval and for YouTube Dataset, 24 frames for data balance. We average the score across all three splits.

Table 1 depicts the comparison between the baselines on UCF101 dataset. We can see that with the naive alternating method, while the model can perform multi-task prediction, the performance cannot outperform single task models. In contrast, the model trained with the proposed method not only is able to predict multi-task prediction of action and caption embedding but also outperforms both of the single-task models. Also, it outperforms both single-task models and the naive alternating method in terms of both action recognition and caption embedding task.

|  | Hit@1 | Accuracy (%) |
|---|---|---|
| Single task | 56.00 | 51.58 |
| DML (w/o LWF) | 54.44 | 51.26 |
| DML + LWF | **59.04** | **52.58** |

Table 2. Action recognition results on HMDB51 dataset. The proposed model outperforms both the model trained only ot the target data (Single task) and naive DML model.

|  | UCF101 | HMDB51 |
|---|---|---|
| CNN [24] | 65.4 | - |
| Spatial stream CNN [41] | 73.0 | 40.5 |
| C3D [43] | 80.9 | 51.7 |
| C3D + DML (w/o LWF) | 79.6 | 51.3 |
| C3D + DML + LWF (**Ours**) | **82.8** | **52.6** |

Table 3. Comparison results on UCF101 dataset and HMDB51 dataset with other methods with respect to average classification accuracy (%). our method shows the best performance among the works using only RGB input and single CNN model with simple mean pooling.

|  | Class | Caption | | | | |
|---|---|---|---|---|---|---|
|  | mAP | R@1 | R@5 | R@10 | Med r | Mean r |
| Class only | 54.01 | - | - | - | - | - |
| Caption only | - | 40.8 | 76.4 | 87.1 | **1.9** | **5.0** |
| Finetuning (caption→class) | 54.16 | 1.0 | 6.0 | 15.3 | 34.4 | 35.0 |
| Finetuning (class→caption) | 2.61 | 39.5 | 77.2 | 87.2 | 2.0 | **5.0** |
| LWF [29] (caption→class) | 54.38 | 39.1 | 73.7 | 85.9 | 2.3 | 5.7 |
| LWF [29] (class→caption) | 52.79 | 40.7 | 76.8 | 87.1 | 2.0 | **5.0** |
| DML | 52.33 | 37.0 | 73.2 | 84.6 | 2.3 | 5.7 |
| DML + LWF (**Ours**) | **54.79** | **40.9** | **77.9** | **87.7** | **1.9** | **5.0** |

Table 4. Comparison results trained on PASCAL VOC2012 and Flickr8k data. The proposed model outperforms both class-only model and caption-only model.

With our method, captioning data can be used for helping the model to learn action recognition task. By adjusting the data weight balancing, our method can be used to improve single-task performance. In this experiment, we use DML for improving action recognition performance on HMDB51 dataset by exploiting YouTube2Text dataset. The comparison of the action recognition performance on HMDB51 dataset is depicted in Table 2. We can see the action recognition performance of the proposed model is improved compared to both single task model and the naive alternating method. We can conclude that by simply adding a few captioning videos, we can achieve performance improvement in action recognition task.

We also compare the model trained by our method with other action recognition models that use RGB frames as input. As a reference, we compare with several previous action recognition works based on CNN models [24, 41]. As we use C3D [43] as a baseline, we follow its same evaluation procedure. We extract 3D CNN activations from the fc6 layer, average pool and L2 normalize them to obtain the video representation. With this video vector, we train a SVM to perform video-level classification.

The comparison is depicted in Table 3. In the top part of the table, we quote the performance scores from [24, 41]. In the bottom part, denoted by C3D, we use the same settings and hyper-parameters as [43] and report the performance.

Among the works using only RGB input and single CNN model, our method shows the best performance in both UCF101 and HMDB51 video action recognition task. We claim this is meaningful because by only using a few captioning videos, we can achieve performance improvement.

### 4.3. Multi-task with Heterogeneous Image Data

In order to verify our method, we also train and test our model on the image domain. For classification, we use PASCAL VOC 2012 [16] and for caption task, we use Flickr 8k [35] dataset. PASCAL VOC 2012 contains 5717 training images with 20 class labels and Flickr 8k dataset has 6000 training images so we think the two datasets can be considered to be balanced. If we see the images in PASCAL VOC classification data [16], the classes are the objects from natural scenes, which Flickr 8k captioning data [35] also deals with. We decide that PASCAL VOC classification dataset is more related to Flickr8k dataset than PASCAL action dataset because image-based models deal with the appearance rather than motion.

The comparison between several baselines is depicted in Table 4. In this experiment, we additionally compare with models which are trained via fine-tuning and the "Learning-without-Forgetting" [29] method, denoted as "Finetuning" and "LWF" respectively. By supervising only with either classification or captioning data, the model is not able to perform the opposite task. By fine-tuning the model that has been pre-trained on the opposite task, the model achieves slightly better performance on the target task. However, due to the forgetting effect, the performance of the source task is poor compared to the model trained only on the source task data. The performance degradation is very large, because of the domain gap of the two heterogeneous datasets. With the LWF method, the model can achieve better performance on the source task compared to the fine-tuned model. Yet, the LWF method fails to outperform our method in terms of the opposite task. As shown in the case of video data, by utilizing alternating optimization, our model shows the best performance among the baselines in both of the tasks. Also, we conclude that the proposed method performs better than the naive alternating method in terms of every metric.

Comparing to the improvement shown in the video experiment, the improvements by the proposed model over the baseline are relatively marginal. We guess that this is relevant to the context of the datasets. In Pascal VOC and

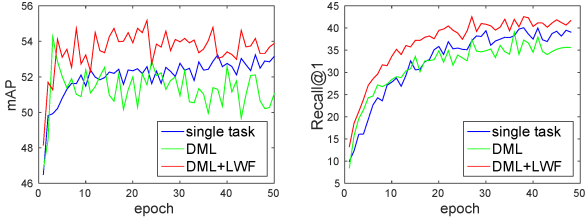

Figure 4. The performance graph through epoch (Class, mAP) and (Caption, Recall@1).

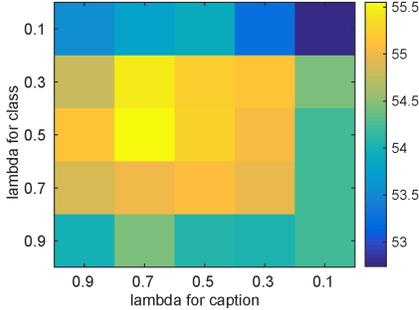

Figure 5. Visualizing of the heat map of mAP score on PASCAL VOC2012 test data with respect to multi-task parameter $\lambda_C$ and $\lambda_S$. The model shows best performance in $\lambda_C = 0.5, \lambda_S = 0.7$.

| Loss | mAP | R@1 | R@5 | R@10 | Med r | Mean r |
|---|---|---|---|---|---|---|
| L1 loss | 55.65 | 28.7 | 65.3 | 81.3 | 3.3 | 6.8 |
| L2 loss | 54.56 | 26.0 | 61.2 | 77.6 | 3.8 | 7.4 |
| Distill (T=3) | 55.21 | 20.0 | 59.3 | 78.6 | 4.1 | 7.5 |
| Distill (T=2) | 55.55 | **31.9** | 70.8 | **85.5** | **2.6** | **5.8** |
| Softmax (T=1) | **56.17** | 31.0 | **71.3** | 84.5 | 3.0 | 5.9 |

Table 5. Comparison of the loss function for distilling activations. Knowledge distillation of $T = 2$ and $T = 1$ (identical to cross entropy loss) show similarly good performance.

Flickr 8k dataset, these images consist of various contexts, such as images about animals or landscape, while the video datasets we used focus on human and its surroundings. In this regards, the image datasets may have a lower probability that co-occurs relevant visual information from both datasets, than the video datasets. Thus, our method would be more effective for the disjoint tasks consisting of contexts highly correlated each other.

In order to prove the effectiveness of LWF for alternating training than the naive method, we illustrate the performance graph through training step of validation classification precision and recall in Figure 4. The red line denotes the proposed method and the green line and the blue line denotes the naive alternating method and the single-task training respectively. With the naive method, it is hard to see the performance improvements than the single-task model. In contrast, with our final model, we show the improvements in terms of both classification and caption performance.

### 4.4. Empirical Analysis

Since training with video data is computationally heavy, we alternatively analyze our method with the image classification and the image-text matching tasks.

**The Multi-task Parameter $\lambda$** In Eq. (10) for our final optimization scenario, we have two tunable parameters $\lambda_C$ and $\lambda_S$. In order to find the best multi-task parameter $\lambda_C$ and $\lambda_S$, we compare among various $\lambda$ pairs and collect mAP for VOC classification. The result is illustrated in Figure 5. We set each of $\lambda$ value from 0.1 to 0.9 with interval 0.2. As shown in the figure, the appearance of the performance with respect to both $\lambda$'s show the concave curve and the best $\lambda$ values with respect to mAP are $\lambda_C = 0.5$ and $\lambda_S = 0.7$. This means rather than setting $\lambda = 1$ (train only with action loss) or setting $\lambda = 0$ (train only with caption loss), there exists mixing parameters that lead to higher performance.

**Choice of Loss Function** In Eq. (7), the value of the temperature $T$ can be also a hyper-parameter. Hinton *et al*. [22] first suggests this objective function and proves that with a high value of T, the function converges to L2 loss function. This means the loss will encourage the network to better encode similarities among classes. Therefore, with this hyperparameter T, we can distill not only the information of top-ranked class but also the distribution of all classes into the network. The author in [29] found that $T = 2$ works best for transferring between image classification tasks according to grid search.

In this experiment, we test $T$ for disjoint multi-task learning between action and caption task. We compare the results of L1, L2 loss, and knowledge distillation loss of several values of $T$. Table 5 illustrates the result of the comparison. We can see that knowledge distillation of $T = 2$ and $T = 1$ (identical to cross entropy loss) show similarly good performance. Among the two, as a final loss function, we choose knowledge distillation with $T = 2$, which achieve the best performance across multiple metrics.

**Qualitative Results** In this section, in order to show the capability of our multi-task model, we demonstrate qualitative results of cross-task prediction, which means the prediction has different modality. Specifically, we test data from action recognition dataset (UCF101) and predict the caption modality, which has not been supervised for the counterpart branch and not been shown to the model.

Figure 6 shows cross-task predictions with UCF101 and YouTube2Text dataset. Figure 7 shows the results with PASCAL VOC classification and Flickr 8k dataset. For caption task, the extracted embeddings are used for searching the nearest sentence from the test split caption dataset.

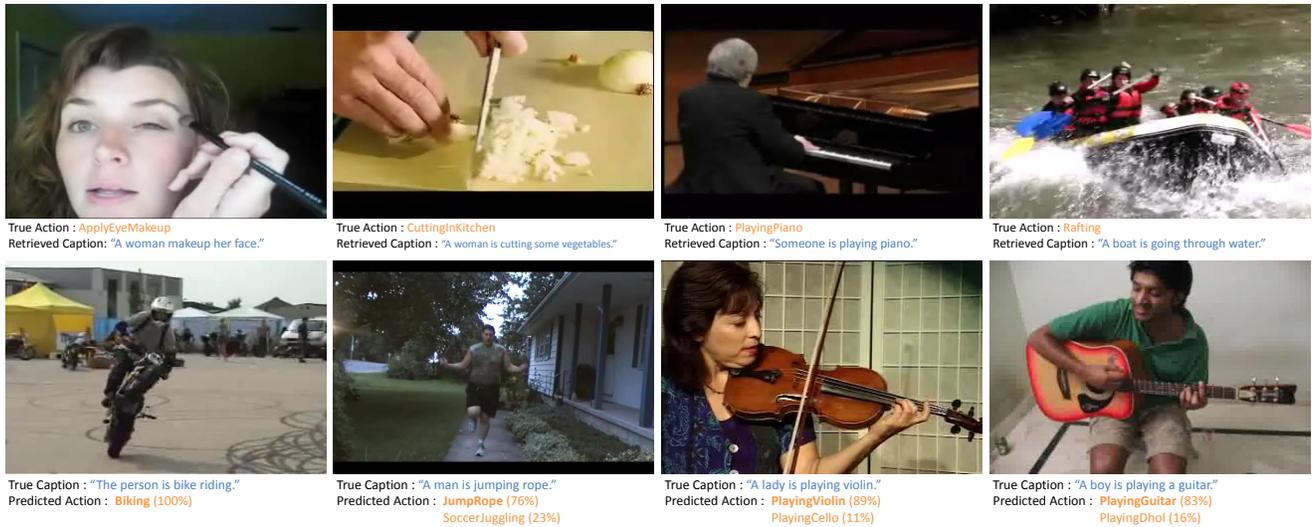

Figure 6. Cross-task prediction results on video datasets. (Top Row : YouTube2Text caption retrieval on UCF101 action data, Bottom Row : UCF101 action prediction with probability on YouTube2Text caption data.)

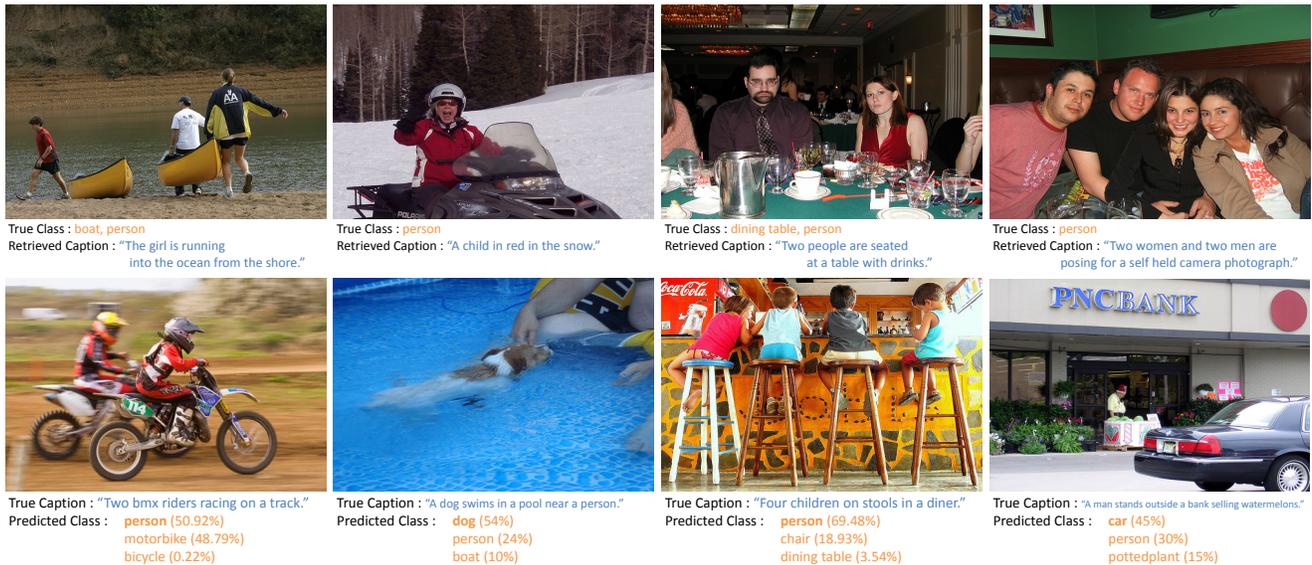

Figure 7. Cross-task prediction results on image datasets. (Top Row : Flickr8k caption retrieval on PASCAL VOC classification data, Bottom Row : PASCAL VOC class prediction with probability on Flickr8k caption data.)

## 5. Conclusion

We have showed disjoint multi-task learning (DML) for human centric tasks, action recognition and caption retrieval. The proposed alternating optimization method with distilling loss shows better performance for both tasks by dealing with the forgetting effect. With these results, we show the possibility to merge various datasets with multiple tasks. From several experiments, we interpret that information from human centric tasks complements each other. Another advantage of our method is that our method is generic; hence we may find other heterogeneous applications that complement each other.

**Acknowledgements.** This work was supported by the Technology Innovation Program (No. 10048320), funded by the Ministry of Trade, Industry & Energy (MI, Korea).

## References

[1] R. Aljundi, P. Chakravarty, and T. Tuytelaars. Expert gate: Lifelong learning with a network of experts. In *The IEEE Conference on Computer Vision and Pattern Recognition (CVPR)*, July 2017. 3

[2] B. D. Argall, S. Chernova, M. Veloso, and B. Browning. A survey of robot learning from demonstration. *Robotics and autonomous systems*, 57(5):469–483, 2009. 1

[3] J. Ba and D. Kingma. Adam: A method for stochastic optimization, 2015. 5

[4] V. Belagiannis and A. Zisserman. Recurrent human pose estimation. *arXiv preprint arXiv:1605.02914*, 2016. 2

[5] H. Bilen and A. Vedaldi. Integrated perception with recurrent multi-task neural networks. In D. D. Lee, M. Sugiyama, U. V. Luxburg, I. Guyon, and R. Garnett, editors, *Advances in Neural Information Processing Systems 29*, pages 235–243. Curran Associates, Inc., 2016. 2

[6] H. Bilen and A. Vedaldi. Universal representations: The missing link between faces, text, planktons, and cat breeds. *arXiv preprint arXiv:1701.07275*, 2017. 2

[7] Z. Cao, T. Simon, S.-E. Wei, and Y. Sheikh. Realtime multi-person 2d pose estimation using part affinity fields. *arXiv preprint arXiv:1611.08050*, 2016. 1

[8] J. Carreira and A. Zisserman. Quo vadis, action recognition? a new model and the kinetics dataset. In *The IEEE Conference on Computer Vision and Pattern Recognition (CVPR)*, July 2017. 2

[9] R. Caruana. Multitask learning. In *Learning to learn*, pages 95–133. Springer, 1998. 2

[10] K. Chatfield, K. Simonyan, A. Vedaldi, and A. Zisserman. Return of the devil in the details: Delving deep into convolutional nets. *arXiv preprint arXiv:1405.3531*, 2014. 5

[11] D. L. Chen and W. B. Dolan. Collecting highly parallel data for paraphrase evaluation. In *Proceedings of the 49th Annual Meeting of the Association for Computational Linguistics: Human Language Technologies-Volume 1*, pages 190–200. Association for Computational Linguistics, 2011. 3, 5

[12] J. Dai, K. He, and J. Sun. Instance-aware semantic segmentation via multi-task network cascades. In *Proceedings of the IEEE Conference on Computer Vision and Pattern Recognition*, pages 3150–3158, 2016. 2

[13] V. Delaitre, I. Laptev, and J. Sivic. Recognizing human actions in still images: a study of bag-of-features and part-based representations. In *BMVC 2010-21st British Machine Vision Conference*, 2010. 1, 2

[14] A. Diba, V. Sharma, and L. Van Gool. Deep temporal linear encoding networks. In *The IEEE Conference on Computer Vision and Pattern Recognition (CVPR)*, July 2017. 2

[15] J. Donahue, L. Anne Hendricks, S. Guadarrama, M. Rohrbach, S. Venugopalan, K. Saenko, and T. Darrell. Long-term recurrent convolutional networks for visual recognition and description. In *Proceedings of the IEEE Conference on Computer Vision and Pattern Recognition*, pages 2625–2634, 2015. 2

[16] M. Everingham, S. A. Eslami, L. Van Gool, C. K. Williams, J. Winn, and A. Zisserman. The pascal visual object classes challenge: A retrospective. *International Journal of Computer Vision*, 111(1):98–136, 2015. 6

[17] R. Girshick. Fast r-cnn. In *Proceedings of the IEEE International Conference on Computer Vision*, pages 1440–1448, 2015. 2

[18] A. Gupta, A. Kembhavi, and L. S. Davis. Observing human-object interactions: Using spatial and functional compatibility for recognition. *IEEE Transactions on Pattern Analysis and Machine Intelligence*, 31(10):1775–1789, 2009. 1, 2

[19] K. He, G. Gkioxari, P. Dollár, and R. Girshick. Mask r-cnn. *arXiv preprint arXiv:1703.06870*, 2017. 1

[20] S. Herath, M. Harandi, and F. Porikli. Going deeper into action recognition: A survey. *Image and Vision Computing*, 60:4–21, 2017. 2

[21] C. Heyes and C. Foster. Motor learning by observation: evidence from a serial reaction time task. *The Quarterly Journal of Experimental Psychology: Section A*, 55(2):593–607, 2002. 1

[22] G. Hinton, O. Vinyals, and J. Dean. Distilling the knowledge in a neural network. *arXiv preprint arXiv:1503.02531*, 2015. 4, 7

[23] A. Kar, N. Rai, K. Sikka, and G. Sharma. Adascan: Adaptive scan pooling in deep convolutional neural networks for human action recognition in videos. In *The IEEE Conference on Computer Vision and Pattern Recognition (CVPR)*, July 2017. 2

[24] A. Karpathy, G. Toderici, S. Shetty, T. Leung, R. Sukthankar, and L. Fei-Fei. Large-scale video classification with convolutional neural networks. In *Proceedings of the IEEE conference on Computer Vision and Pattern Recognition*, pages 1725–1732, 2014. 1, 2, 5, 6

[25] J. Kirkpatrick, R. Pascanu, N. Rabinowitz, J. Veness, G. Desjardins, A. A. Rusu, K. Milan, J. Quan, T. Ramalho, A. Grabska-Barwinska, et al. Overcoming catastrophic forgetting in neural networks. *Proceedings of the National Academy of Sciences*, page 201611835, 2017. 3

[26] R. Kiros, R. Salakhutdinov, and R. S. Zemel. Multimodal neural language models. In *ICML*, volume 14, pages 595–603, 2014. 2

[27] I. Kokkinos. Ubernet: Training a universal convolutional neural network for low-, mid-, and high-level vision using diverse datasets and limited memory. In *The IEEE Conference on Computer Vision and Pattern Recognition (CVPR)*, July 2017. 2

[28] H. Kuehne, H. Jhuang, E. Garrote, T. Poggio, and T. Serre. Hmdb: a large video database for human motion recognition. In *Computer Vision (ICCV), 2011 IEEE International Conference on*, pages 2556–2563. IEEE, 2011. 5

[29] Z. Li and D. Hoiem. Learning without forgetting. In *European Conference on Computer Vision*, pages 614–629. Springer, 2016. 2, 3, 4, 6, 7

[30] A.-A. Liu, Y.-T. Su, W.-Z. Nie, and M. Kankanhalli. Hierarchical clustering multi-task learning for joint human action grouping and recognition. *IEEE transactions on pattern analysis and machine intelligence*, 39(1):102–114, 2017. 2

[31] M.-T. Luong, Q. V. Le, I. Sutskever, O. Vinyals, and L. Kaiser. Multi-task sequence to sequence learning. *arXiv preprint arXiv:1511.06114*, 2015. 2

[32] B. Mahasseni and S. Todorovic. Regularizing long short term memory with 3d human-skeleton sequences for action recognition. 2

[33] D. T. Nguyen, W. Li, and P. O. Ogunbona. Human detection from images and videos: A survey. *Pattern Recognition*, 51:148–175, 2016. 1

[34] M. Oquab, L. Bottou, I. Laptev, and J. Sivic. Learning and transferring mid-level image representations using convolu-

tional neural networks. In *Proceedings of the IEEE conference on computer vision and pattern recognition*, pages 1717–1724, 2014. 2

[35] C. Rashtchian, P. Young, M. Hodosh, and J. Hockenmaier. Collecting image annotations using amazon's mechanical turk. In *Proceedings of the NAACL HLT 2010 Workshop on Creating Speech and Language Data with Amazon's Mechanical Turk*, pages 139–147. Association for Computational Linguistics, 2010. 6

[36] O. Russakovsky, J. Deng, H. Su, J. Krause, S. Satheesh, S. Ma, Z. Huang, A. Karpathy, A. Khosla, M. Bernstein, et al. Imagenet large scale visual recognition challenge. *International Journal of Computer Vision*, 115(3):211–252, 2015. 2, 5

[37] R. Salakhutdinov, J. B. Tenenbaum, and A. Torralba. Learning with hierarchical-deep models. *IEEE transactions on pattern analysis and machine intelligence*, 35(8):1958–1971, 2013. 1

[38] S. Schaal. Learning from demonstration. In *Advances in neural information processing systems*, pages 1040–1046, 1997. 1

[39] H. Shin, J. K. Lee, J. Kim, and J. Kim. Continual learning with deep generative replay. *arXiv preprint arXiv:1705.08690*, 2017. 3

[40] D. L. Silver, Q. Yang, and L. Li. Lifelong machine learning systems: Beyond learning algorithms. In *AAAI Spring Symposium: Lifelong Machine Learning*, volume 13, page 05, 2013. 3

[41] K. Simonyan and A. Zisserman. Two-stream convolutional networks for action recognition in videos. In *Advances in Neural Information Processing Systems*, pages 568–576, 2014. 1, 2, 6

[42] K. Soomro, A. R. Zamir, and M. Shah. Ucf101: A dataset of 101 human actions classes from videos in the wild. *arXiv preprint arXiv:1212.0402*, 2012. 3, 5

[43] D. Tran, L. Bourdev, R. Fergus, L. Torresani, and M. Paluri. Learning spatiotemporal features with 3d convolutional networks. In *2015 IEEE International Conference on Computer Vision (ICCV)*, pages 4489–4497. IEEE, 2015. 1, 2, 5, 6

[44] I. Vendrov, R. Kiros, S. Fidler, and R. Urtasun. Order-embeddings of images and language. *arXiv preprint arXiv:1511.06361*, 2015. 3, 5

[45] O. Vinyals, A. Toshev, S. Bengio, and D. Erhan. Show and tell: A neural image caption generator. In *Proceedings of the IEEE Conference on Computer Vision and Pattern Recognition*, pages 3156–3164, 2015. 2

[46] J. Wang and J. Ye. Safe screening for multi-task feature learning with multiple data matrices. In *International Conference on Machine Learning*, pages 1747–1756, 2015. 2

[47] L. Wang, Y. Xiong, Z. Wang, and Y. Qiao. Towards good practices for very deep two-stream convnets. *arXiv preprint arXiv:1507.02159*, 2015. 2

[48] X. Wang, A. Farhadi, and A. Gupta. Actions˜ transformations. *arXiv preprint arXiv:1512.00795*, 2015. 2

[49] S.-E. Wei, V. Ramakrishna, T. Kanade, and Y. Sheikh. Convolutional pose machines. In *Proceedings of the IEEE Conference on Computer Vision and Pattern Recognition*, pages 4724–4732, 2016. 1

[50] Q. Wu, C. Shen, L. Liu, A. Dick, and A. van den Hengel. What value do explicit high level concepts have in vision to language problems? In *Proceedings of the IEEE Conference on Computer Vision and Pattern Recognition*, pages 203–212, 2016. 2

[51] B. Yao and L. Fei-Fei. Modeling mutual context of object and human pose in human-object interaction activities. In *Computer Vision and Pattern Recognition (CVPR), 2010 IEEE Conference on*, pages 17–24. IEEE, 2010. 1, 2

[52] Q. You, H. Jin, Z. Wang, C. Fang, and J. Luo. Image captioning with semantic attention. In *Proceedings of the IEEE Conference on Computer Vision and Pattern Recognition*, pages 4651–4659, 2016. 2

[53] L. Yu, E. Park, A. C. Berg, and T. L. Berg. Visual madlibs: Fill in the blank image generation and question answering. *arXiv preprint arXiv:1506.00278*, 2015. 1

[54] T. Zhang, B. Ghanem, S. Liu, and N. Ahuja. Robust visual tracking via multi-task sparse learning. In *Computer vision and pattern recognition (CVPR), 2012 IEEE conference on*, pages 2042–2049. IEEE, 2012. 2

[55] Z. Zhang, P. Luo, C. C. Loy, and X. Tang. Facial landmark detection by deep multi-task learning. In *European Conference on Computer Vision*, pages 94–108. Springer, 2014. 2